\def\year{2019}\relax
\documentclass[letterpaper]{article} 
\usepackage{aaai19}  
\usepackage{times}  
\usepackage{helvet}  
\usepackage{courier}  
\usepackage{url}  
\usepackage{graphicx}  
\usepackage{multirow}
\usepackage{amsmath}
\usepackage{booktabs}
\begin{document}
%
\title{Spatial and Temporal Mutual Promotion for Video-based Person Re-identification}
\author{Yiheng Liu\(^1\),\quad Zhenxun Yuan\(^2\),\quad Wengang Zhou\(^1\),\quad Houqiang Li\(^1\)\\
\(^1\)CAS Key Laboratory of Technology in GIPAS, EEIS Department, University of Science and Technology of China \\ \(^2\) School of Electrical and Computer Engineering, Purdue University\\
lyh156@mail.ustc.edu.cn,\quad yuan141@purdue.edu,\quad \{zhwg,lihq\}@ustc.edu.cn\\
}
\maketitle
\begin{abstract}
Video-based person re-identification is a crucial task of matching video sequences of a person across multiple camera views. Generally, features directly extracted from a single frame suffer from occlusion, blur, illumination and posture changes. This leads to false activation or missing activation in some regions, which corrupts the appearance and motion representation. How to explore the abundant spatial-temporal information in video sequences is the key to solve this problem. To this end, we propose a Refining Recurrent Unit (RRU) that recovers the missing parts and suppresses noisy parts of the current frame's features by referring historical frames. With RRU, the quality of each frame's appearance representation is improved. Then we use the Spatial-Temporal clues Integration Module (STIM) to mine the spatial-temporal information from those upgraded features. Meanwhile, the multi-level training objective is used to enhance the capability of RRU and STIM. Through the cooperation of those modules, the spatial and temporal features mutually promote each other and the final spatial-temporal feature representation is more discriminative and robust. Extensive experiments are conducted on three challenging datasets, \emph{i.e.,} iLIDS-VID, PRID-2011 and MARS. The experimental results demonstrate that our approach outperforms existing state-of-the-art methods of video-based person re-identification on iLIDS-VID and MARS and achieves favorable results on PRID-2011.
\end{abstract}

\section{Introduction}
Person re-identification aims to identify persons across different cameras views.
Recently, this topic has drawn more and more attention thanks to its significant applications in video surveillance analysis and retrieval.
This task is very challenging due to background clutter, blur, occlusion, as well as the dramatic variation in illumination, pedestrian's postures and viewpoints.
Generally, person re-identification is approached with either image or video data for representation.
Many image-based person re-identification methods have achieved impressive progress.
However, those methods are susceptible to the quality of images.
Limited amount of information in a single image results in a lower tolerance to noise.
For similar pedestrians, if the discriminative patches are lost due to occlusion or blur, it will easily lead to misidentification.

Compared with a single image to represent a person, a video sequence contains richer information \cite{zheng2016mars}.
The information of different video frames complements each other, so that it is more robust to noise.
More importantly, motion context in video sequences is useful for identifying pedestrians.
Therefore, how to make full use of spatial-temporal information is the key to video-based person re-identification.
Some methods \cite{mclaughlin2016recurrent,zhou2017see,xu2017jointly} use recurrent networks to fuse temporal information.
Another alternative is to predict a quality score for whole or part of each video frame \cite{liu2017quality,song2017region,li2018diversity}.
However, a key effect of temporal information is ignored by these methods: the temporal information is useful to resist spatial noise.
Usually, in a video sequence, although a region in one frame is corrupted by noise, the missing information can be recovered by regions of the same location in some other frames.
Given a video frame, the difficulty lies in how to refine the noisy regions with the features of previous video frames.

In this paper, we propose a new approach to handle the difficulty of video-based person re-identification.
Instead of directly using the features of each video frame to extract temporal features,
we first propose a refining recurrent unit (RRU) to recover the missing parts
and suppress noisy parts based on the appearance and motion context from historical video frames.
After that, with the refined feature representation, we introduce a spatial-temporal clues integration module (STIM) to integrate spatial information and temporal information simultaneously.
Meanwhile, the proposed multi-level training objective further enhances the capability of RRU and STIM.
The cooperation of those modules enables the network to learn more robust and discriminative feature representation for accurate person re-identification.

\section{Related Works}
Most prior works on person re-identification  are based on still images and dedicated to two key issues: discriminative feature representation learning and distance metric learning.
Features that are both discriminative and invariant to background and viewpoint changes are crucial to person re-identification.
Handcrafted features such as color histograms \cite{karanam2015person}, texture histograms \cite{gray2008viewpoint} and Local Binary Pattern \cite{xiong2014person} are widely utilized.
With feature representation, distance metric learning approaches are widely explored to accurately measure the similarity between pedestrians.
Following such a paradigm, many effective methods have been proposed, such as LADF \cite{li2013learning}, RankSVM \cite{zhao2014learning}, XQDA \cite{liao2015person}.

In recent years, the rapidly developing Convolutional Neural Networks (CNN) have greatly advanced the progress of person re-identification.
Subramaniam \emph{et al}. compute the normalized correlation between two patch matrices to handle inexact matching problems \cite{subramaniam2016deep}.
Wang \emph{et al}. propose an approach to learn single image representation and cross-image representation simultaneously \cite{wang2016joint}.
In \cite{qian2017multi}, a multi-scale and saliency-based model is proposed to learn deep discriminative feature representations at different scales.
Si \emph{et al}. design a dual attention network to learn context-aware feature sequences and apply dually attentive comparison \cite{si2018dual}.
Meanwhile, many works try to improve the representation capability of models based on local parts.
Wei \emph{et al}. use a part extraction module to generate part regions and learn discriminative features based on different part regions \cite{wei2017glad}.
Sun \emph{et al}. propose a refined part pooling to enhance the consistency in each part \cite{sun2017beyond}.
Zhao \emph{et al}. use a pretrained human landmark generation model to extract body regions
and then use a tree-structured fusion strategy to integrate the full-body features and different body subregion features \cite{zhao2017spindle}.

On the other hand, video-based person re-identification has also gained considerable attention.
Wang \emph{et al}. propose a model that automatically selects more discriminative video fragments from the whole video sequence and learns cross-view matching by ranking \cite{wang2014person}.
You \emph{et al}. design a top-push distance learning model to enlarge the inter-class margin and minimize intra-class variations in order to improve the matching accuracy \cite{you2016top}.
Recently, some other methods \cite{liu2017quality,song2017region,li2018diversity} have been proposed to predict the quality scores for the features of video frames or local regions.
They average the frame or region features in a weighting manner based on the quality scores as the final representation of a video sequence.
Although their strategies alleviate the noise problem, the neglect of important temporal context limits their capability.
To make better use of the temporal information in video sequences,
RNN or LSTM \cite{vinyals2015show} is adopted to fuse the feature vectors extracted from video frames \cite{mclaughlin2016recurrent,zhou2017see}.
The temporal average pooling \cite{mclaughlin2016recurrent} is applied at each time step to generate the final representation of video sequences.
Although the temporal clues are captured with the recurrent models, the motion context of different regions in video sequences is lost, which results in limited performance.

Different from the above methods, our work aims to take full advantage of spatial-temporal information in video sequences. To this end, we propose two effective modules, \emph{i.e.,} refining recurrent unit (RRU) and spatial-temporal clues integration module (STIM). The former resists noise and recovers missing activation occurred in different regions while the latter integrates both spatial and temporal clues in a unified framework. Further, we apply a multi-level training objective to simultaneously optimize them. To the best of our knowledge, this is the first attempt to refine appearance representation with spatial-temporal information in person re-identification task which achieves promising performance.


\begin{figure*}[t]
  \centering
  \includegraphics[width=15cm]{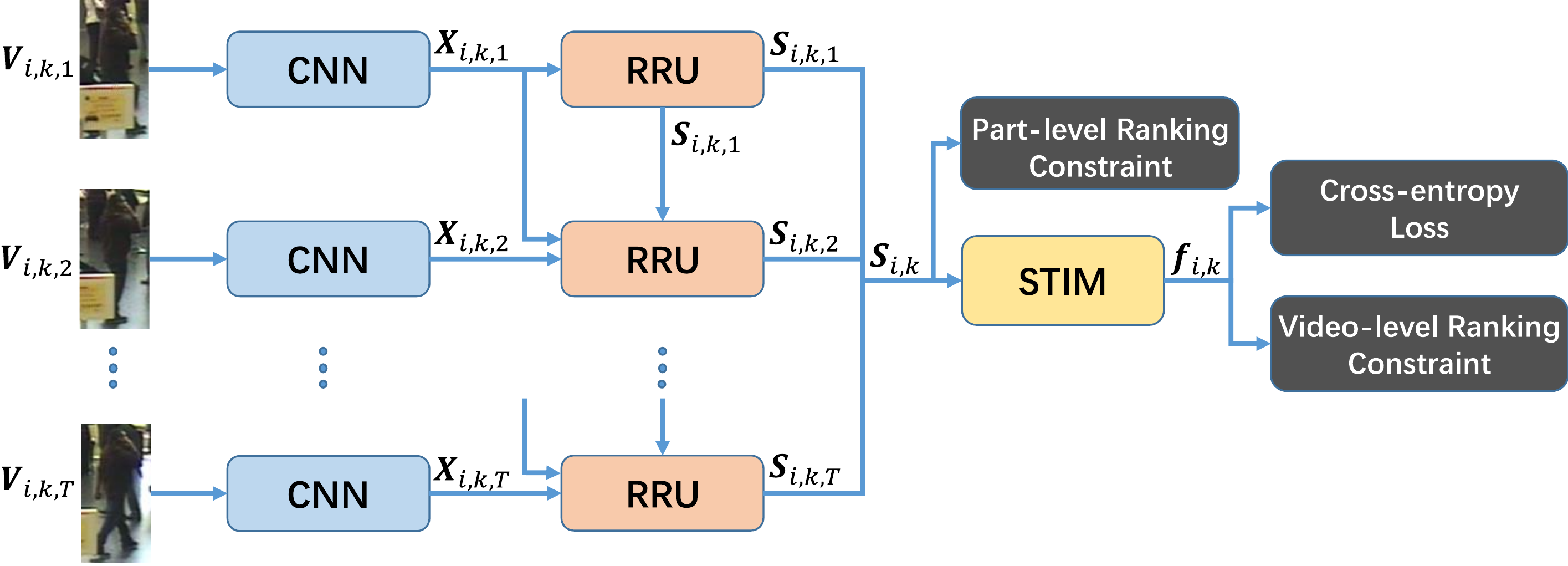}
  \caption{The overall architecture of the proposed method. The weights of all layers are shared for each time step. RRU: refining recurrent unit. STIM: spatial-temporal clues integration module. The auxiliary classifier is not drawn for the convenience of display.}
\end{figure*}

\section{Method}
In this section, we first introduce the overall architecture of the proposed method.
Then we elaborate the key components in our framework, including refining recurrent unit (RRU), spatial-temporal clues integration module (STIM) and multi-level training objective, separately.

\subsection{Framework Overview}
The overall architecture of the proposed model is shown in Fig.~1.
\(\textbf{V}_i = \{\textbf{V}_{i,k}\}_{k=1}^{K}\) represents \(K\) video sequences of person \(i\).
Each video sequence contains \(T\) frames.
We use \(\textbf{V}_{i, k, t}\) to represent the \(t^{th}\) frame of \(\textbf{V}_{i, k}\).
We adopt Inception-v3 \cite{szegedy2016rethinking} as the backbone of feature extraction module.
Given a video sequence \(\textbf{V}_{i, k}\), each frame \(\textbf{V}_{i, k, t}\) is fed to Inception-v3 module
to extract frame-level feature maps \(\textbf{X}_{i, k, t} \in {\cal{R}}^{C \times H \times W}\) from the outputs of the final inception block.
Feature maps of all frames in a video sequence are fed into RRU for refinement.
The refined feature maps are then fed into STIM to generate the final video-level feature representation
\(\textbf{f}_{i,k} \in {\cal{R}}^{256}\).

\subsection{Refining Recurrent Unit}
In the involved video data for person re-identification, it is common that some regions of a target person suffer from occlusion, blur and varied postures.
In the degraded regions, the raw feature maps \(\textbf{X}_{i, k, t}\) are easily polluted.
However, the appearance and motion information remembered from regions of the same position in other frames can help to recover the lost information.
Inspired by such observation, we design a recurrent unit to remove noise and recover missing activation regions by implicitly referring the appearance and motion information extracted from previous video frames.

As illustrated in Fig.~2(a), for time step \(t\), RRU takes three inputs: the raw feature \(\textbf{X}_{i, k, t-1}\) and the refined feature \(\textbf{S}_{i, k, t-1}\) of the last frame,
as well as the raw feature \(\textbf{X}_{i, k, t}\) of current frame.
In the first time step, \(\textbf{X}_{i, k, 0}\) and \(\textbf{S}_{i, k, 0}\) are initialized with \(\textbf{X}_{i, k, 1}\).
The update gate model \(g\) contained in RRU is used to decide how to update \(\textbf{S}_{i, k, t}\).

The refined appearance representation \(\textbf{S}_{i, k, t-1}\) of last frame lends itself as a good reference to evaluate the quality of current frame.
So we use \((\textbf{X}_{i, k, t} - \textbf{S}_{i, k, t-1})\) as an input of \(g\) to model the appearance differences.
Between neighboring frames, pedestrian's movement causes translation of feature responses in spatial space.
With such consideration, we use \((\textbf{X}_{i, k, t}- \textbf{X}_{i, k, t-1})\) as another input of \(g\) to capture motion context.
These two terms are concatenated to form \(\textbf{Z}_{i} \in {\cal{R}}^{2C \times H \times W}\) for \(g\) to determine the update gate of different regions.
The update gate \(\textbf{Z}\) is defined as follows,
\begin{equation}
 \textbf{Z} = g(\textbf{Z}_{i}) = g( [\textbf{X}_{i, k, t}- \textbf{S}_{i, k, t-1}, \textbf{X}_{i, k, t} - \textbf{X}_{i, k, t-1}])\,.
\end{equation}

Fig.~2(b) illustrates the detailed structure of update gate model \(g\).
The first layer of the update gate model \(g\) named transition layer consists of a convolutional layer with 256 filters of size \(1 \times 1\), a batch normalization (BN) layer and a rectified linear unit (ReLU).
It is designed to summarize the appearance and motion information of each spatial location and reduce the feature dimension.
Then a spatial attention model and a channel attention model are applied to the transitional feature \(\textbf{Z}_{t} \in {\cal{R}}^{256 \times H \times W}\) produced by the first layer, separately.

For spatial attention model, we first use a global cross-channel average pooling layer to get the overall response in each spatial position.
Then two FC layers (the first 128-node FC layer is followed by ReLU) are applied to generate the spatial attention maps \(\textbf{Z}_{s} \in {\cal{R}}^{1 \times H \times W}\).
It is formulated as
\begin{equation}
 \textbf{Z}_{s} = \textbf{W}^2_s \times ReLU(\textbf{W}^1_s \times \textbf{Z}^{H,W}_t) \,,
\end{equation}
where \(\textbf{Z}^{H,W}_t \in {\cal{R}}^{H, W}\) is the result of \(\textbf{Z}_t\) after cross-channel average pooling and \(\times\) is matrix multiplication.

For channel attention model,  we first introduce a global spatial space average pooling layer to get overall response of each channel.
Then a FC layer is applied to get the channel attention maps \(\textbf{Z}_{c} \in {\cal{R}}^{C \times 1 \times 1}\), which is formulated as
\begin{equation}
 \textbf{Z}_{c} = \textbf{W}_c \times \textbf{Z}^{C}_t \,,
\end{equation}
where \(\textbf{Z}^C_t \in {\cal{R}}^{C}\) is the result of \(\textbf{Z}_t\) after global spatial space average pooling.

The overall attention maps of current input feature are the product of spatial attention maps and channel attention maps.
After a sigmoid operation, the overall attention maps are normalized into the range between 0 and 1, formulated as
\begin{equation}
 \textbf{Z} = sigmoid(\textbf{Z}_{s} \odot \textbf{Z}_{c}) \,,
\end{equation}
where \(\textbf{Z} \in {\cal{R}}^{C \times H \times W}\) is the update gate of \(\textbf{X}_{i, k, t}\) and \(\odot\) denotes element-wise multiplication.
Then RRU refines \(\textbf{X}_{i, k, t}\) with the previous refined feature \(\textbf{S}_{i,k,t-1}\) as
\begin{equation}
 \textbf{S}_{i, k, t} = (1 - \textbf{Z}) \odot \textbf{S}_{i, k, t-1} + \textbf{Z} \odot \textbf{X}_{i, k, t} \,,
\end{equation}
where \(\textbf{S}_{i, k, t} \in {\cal{R}}^{C \times H \times W}\) is the refined feature of the raw input \(\textbf{X}_{i, k, t}\) of the current video frame.
The value of each position in \(\textbf{Z}\) denotes the probability for the activation value in corresponding position of \(\textbf{X}_{i, k, t}\) to be reserved.
Higher probability value indicates that the update gate model \(g\) considers that feature in this location has high quality and should be reserved,
and on the contrary, locations with lower probability will be dominated by previous refined feature.
Given the refined feature \(\textbf{S}_{i, k, t}\) of each frame in a video sequence, we stack them to \(\textbf{S}_{i, k} \in {\cal{R}}^{C \times T \times H \times W}\), where \(T\) is the number of frames in this video sequence.

Unlike the previous recurrent units (RNN or LSTM), RRU is used to upgrade the frame-level feature by referring the spatial-temporal information instead of extracting new features from temporal feature vectors.
From the above formulas, we can see that \(\textbf{S}_{i, k, t}\) and \(\textbf{X}_{i, k, t}\) share the same feature space, which means that RRU refines features in the same feature space.
This allows RRU to be applied to other video-based models to reduce spatial noise and improve quality of feature at each time step.
We will provide justification on this issue in the later experiments section.

\begin{figure*}[t]
  \centering
  \begin{minipage}[t]{0.48\textwidth}
  \centering
  \includegraphics[width=9cm]{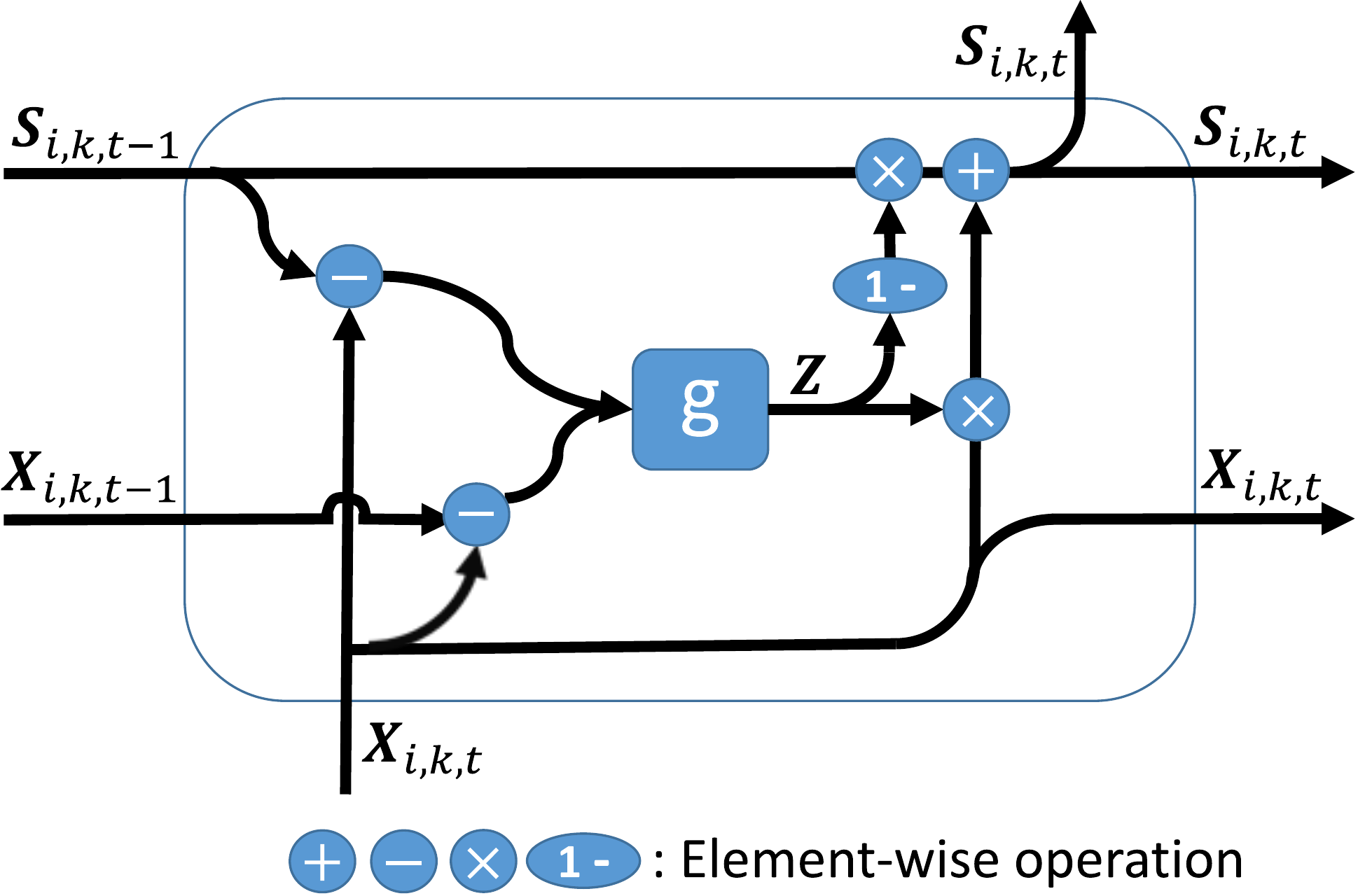}
  \centerline{(a)}
  \end{minipage}
  \begin{minipage}[t]{0.48\textwidth}
  \centering
  \includegraphics[width=7.8cm]{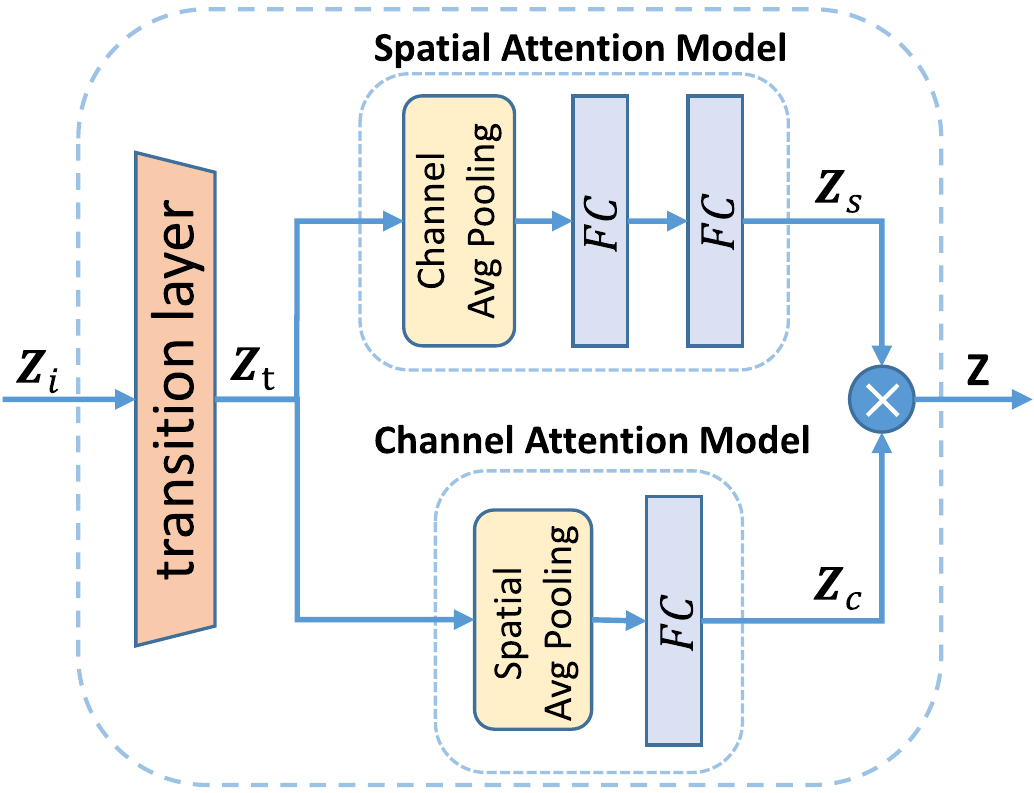}
  \centerline{(b)}
  \end{minipage}
  \caption{(a) The architecture of refining recurrent unit (RRU). \(g\) is the update gate model. (b) The architecture of proposed update gate model \(g\).}
\end{figure*}

\subsection{Spatial-temporal Clues Integration Module}
Many previous works use RNN or weighted average pooling to integrate temporal information from feature vectors that are extracted from video frames.
In those methods, the motion context in different spatial locations is ignored \cite{zhou2017see,liu2017quality}.
In order to fully exploit spatial-temporal information, we propose a spatial-temporal clues integration module (STIM) to
enable our model to learn the appearance representation and capture motion context simultaneously.

STIM contains two 3D convolution blocks and a global average pooling layer.
Each 3D convolution block is composed of three consecutive operations: a 3D convolutional layer \cite{tran2015learning} with 256 filters followed by a 3D BN layer and ReLU.
In the first 3D convolution block, the kernel size is set to be \(1 \times 1 \times 1\), so as to reduce the feature dimension.
The second 3D convolution block with \(3 \times 3 \times 3\) kernels outputs the spatial-temporal feature maps \(\textbf{O}_{i,k} \in {\cal{R}}^{256 \times T \times H \times W}\).
After temporal and spatial space average pooling, we get the final feature representation \(\textbf{f}_{i,k} \in {\cal{R}}^{256}\) as
\begin{equation}
 \textbf{f}_{i,k} = \frac{1}{T}\frac{1}{H}\frac{1}{W}\sum_{t=1}^{T}\sum_{h=1}^{H}\sum_{w=1}^{W}\textbf{O}_{i,k,t} \,.
\end{equation}

The \(3 \times 3 \times 3\) convolution kernels allows STIM to capture the movement of human body parts in spatial space.
Meanwhile, the simultaneous integration of spatial information and temporal information also helps STIM to resist local spatial noise. 

\subsection{Multi-level Training Objective}
The identity classifier and auxiliary classifier of raw Inception-v3 model \cite{szegedy2016rethinking} are retained to learn more robust and informative features.
These two cross-entropy losses are denoted as \({\cal{L}}_c\).
We replace the FC layer in identity classifier with a classifier block \cite{zhong2017camera},
which consists of a 512-node FC layer, BN, ReLU, Dropout, another FC layer and a Softmax layer.
We feed \(\textbf{f}_{i,k}\) to classifier block to generate the prediction of identity.

In addition to the cross-entropy losses, we propose a multi-level training objective to further enhance the capability of RRU and STIM,
which consists of the video-level ranking constraint \({\cal{L}}_v\) and part-level ranking constraint \({\cal{L}}_p\).
Then, the overall training objective is defined as
\begin{equation}
 {\cal{L}} = {\cal{L}}_c + {\cal{L}}_v + {\cal{L}}_p \,.
\end{equation}

We apply batch hard triplet loss \cite{hermans2017defense} on \(\textbf{f}_{i,k}\) as the video-level ranking constraint, formulated as
\begin{equation}
\begin{split}
 {\cal{L}}_v = \frac{1}{N}\frac{1}{K}\sum_{i=1}^{N}\sum_{a=1}^{K}[m &+ \max_{p=1 \cdots K}D(\textbf{f}_{i,a}, \textbf{f}_{i,p}) \\
 &- \min_{\substack{j=1 \cdots N \\ n=1 \cdots K \\ j \neq i}}D(\textbf{f}_{i,a},\textbf{f}_{j,n}))]_+ \,,
\end{split}
\end{equation}
where \(m\) is the margin and \(D(\cdot , \cdot)\) denotes the Euclidean distance between two feature vectors.
\(N\) denotes the number of pedestrians in a mini-batch.
The video-level ranking constraint forces the distance between the overall representation of positive pairs to be smaller than negative pairs.

However, as the overall description of a video sequence, \(\textbf{f}_{i,k}\) ignores the spatial differences.
The most important function of RRU is to refine features of different regions.
So the video-level ranking constraint can not let RRU exert its full capability.
To further enhance the refining capability of RRU in local parts,
we propose the part-level ranking constraint based on the batch hard triplet loss.
We split the original feature maps \(\textbf{S}_{i,k,t}\) output by RRU into \(H\) horizontal strips and get the part-level feature representation as follows,
\begin{equation}
 \textbf{p}^{r}_{i,k} = \frac{1}{T}\frac{1}{W}\sum_{t=1}^{T}\sum_{w=1}^{W}\textbf{S}^{r}_{i,k,t} \,,
\end{equation}
where \(\textbf{S}^{r}_{i,k,t} \in {\cal{R}}^{C \times W}\) is the strip feature from \(r^{th}\) row of \(\textbf{S}_{i,k,t}\).
Given the part-level feature vector \(\textbf{p}^{r}_{i, k} \in {\cal{R}}^{C}\), we define the part-level ranking constraint as
\begin{equation}
\begin{split}
 {\cal{L}}_p = \frac{1}{N}\frac{1}{K}\frac{1}{H}\sum_{i=1}^{N}\sum_{a=1}^{K}\sum_{r=1}^{H}[m &+ \max_{p=1 \cdots K}D(\textbf{p}^{r}_{i,a}, \textbf{p}^{r}_{i,p}) \\
 &- \min_{\substack{j=1 \cdots N \\ n=1 \cdots K \\ j \neq i}}D(\textbf{p}^{r}_{i,a},\textbf{p}^{r}_{j,n}))]_+ \,.
\end{split}
\end{equation}

\begin{figure*}[t]
  \centering
  \includegraphics[width=17cm]{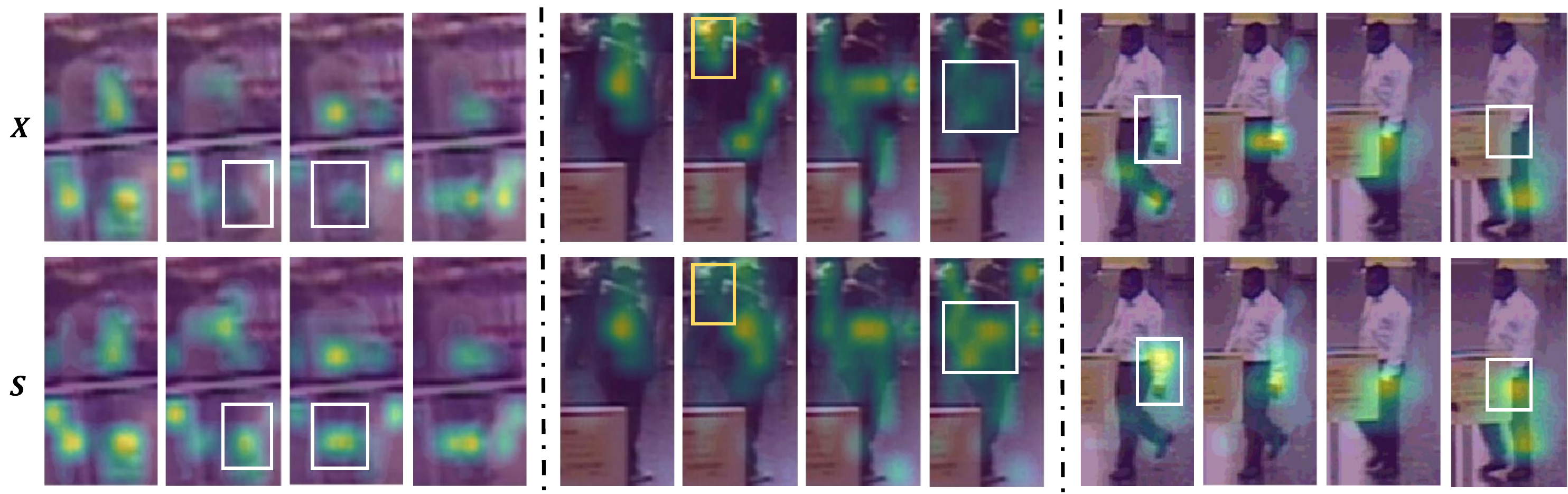}
  \caption{Examples of the feature maps from sequentially sampled frames in a video. The feature responses are shown above the original frame. The first row displays the raw input feature maps \(\textbf{X}\) and the second row shows the refined feature maps \(\textbf{S}\) by RRU. The feature maps are from three different videos which are separated by the dotted line.  }
\end{figure*}

Unlike previous methods \cite{zhao2017spindle,wei2017glad,sun2017beyond}, our part-based method does not require additional parameters, which makes our strategy more efficient and flexible.
For the same strip, part-level ranking constraint pulls features of video sequence belonging to the same identity closer and pushes features of video sequence belonging to different identities farther.
This forces RRU to learn a better update gate model \(g\), which can measure the quality of different regions better.

In our approach, those modules are optimized in a collaborative way.
Given the raw low-quality feature maps of each frame, RRU first refines them based on spatial-temporal information.
Then STIM extracts better spatial-temporal feature representation from high-quality appearance feature maps generated by RRU.
The video-level ranking constraint forces STIM to learn more discriminative spatial-temporal representation for each video sequence.
The part-level ranking constraint helps RRU to focus on different local parts and enhance its refining capability for spatial space.
Through the cooperation between these modules, the final representation is more discriminative and robust to occlusion, blur, illumination and posture changes.

\begin{table*}[ht]
  \caption{Ablation study on each module of the proposed methods on iLIDS-VID and PRID-2011 datasets. The CMC scores (\%) at rank 1, 5, 20 are reported.
  The baseline approach contains only the Inception-v3 model trained by \({\cal{L}}_c\).
  RRU(s) means that we only use the spatial attention \(\textbf{Z}_{s}\) to compute the update gate \(\textbf{Z}\). RRU(c) means that only channel attention \(\textbf{Z}_{c}\) is used. RRU(ad) means that we only use the appearance differences \([\textbf{X}_{i, k, t}- \textbf{S}_{i, k, t-1}]\) as the input of update gate model \(g\).
  RRU(od) denotes that we use \([\textbf{X}_{i, k, t}, \textbf{S}_{i, k, t-1}]\) as the input of update gate model \(g\). When LSTM is inserted into the network, the frame-level feature maps are first fed into a global average pooling layer to get a \(C\)-dim feature vector. The feature vectors of all frames are orderly fed into the LSTM with a hidden size of 256. The temporal average pooling \cite{mclaughlin2016recurrent} is applied to the outputs of LSTM to get the final representation.}
  \centering
  \begin{tabular}{lcccccc}
    \toprule
    \multirow{2}{*}{Method} & \multicolumn{3}{c}{iLIDS-VID} & \multicolumn{3}{c}{PRID-2011} \\
    \cmidrule(r){2-4} \cmidrule(r){5-7}
    & R1 & R5 & R20 & R1 & R5 & R20\\
    \midrule
    Baseline                              & 57.7 & 78.9 & 91.3      & 84.4 & 96.2 & 99.5\\
    Baseline+\({\cal{L}}_v\)                      & 72.2 & 89.7 & 97.5      & 88.0 & 97.2 & 99.9\\
    Baseline+LSTM+\({\cal{L}}_v\)                 & 71.5 & 90.4 & 96.5      & 86.7 & 97.1 & 99.8\\
    Baseline+STIM+\({\cal{L}}_v\)                 & 77.3 & 92.8 & 97.7      & 91.1 & 98.6 & 100.0\\
    \midrule
    Baseline+RRU(s)+\({\cal{L}}_v\)             & 73.5 & 90.4 & 97.0      & 90.8 & 98.5 & 99.9\\
    Baseline+RRU(c)+\({\cal{L}}_v\)             & 73.4 & 90.0 & 96.0     & 88.7 & 97.6 & 99.8\\
    Baseline+RRU(ad)+\({\cal{L}}_v\)             & 74.0 & 89.8 & 96.7      & 90.1 & 97.6 & 99.9\\
    Baseline+RRU(od)+\({\cal{L}}_v\)             & 73.3 & 90.4 & 95.5     & 89.2 & 98.5 & 100.0\\
    Baseline+RRU+\({\cal{L}}_v\)                 & 75.0 & 90.9 & 97.0      & 91.3 & 98.3 & 99.9\\
    \midrule
    Baseline+RRU+LSTM+\({\cal{L}}_v\)            & 72.3 & 90.1 & 97.1      & 89.6 & 98.0 & 100.0\\
    Baseline+RRU+STIM+\({\cal{L}}_v\)            & 80.5 & 94.1 & 98.8      & 91.5 & 98.8 & 99.9\\
    Baseline+RRU+STIM+\({\cal{L}}_{v}\)+\({\cal{L}}_{p}\)    & 84.3 & 96.8 & 99.5      & 92.7 & 98.8 & 99.8\\
    \bottomrule
  \end{tabular}
  \label{model_ablation}
\end{table*}

\begin{table}[ht]
  \caption{Performance comparison for whether RRU or part-level constraint \({\cal{L}}_p\) is added to the network on MARS dataset.}
  \centering
  \begin{tabular}{lccccc}
    \toprule
    Method & mAP & R1 & R5 & R20\\
    \midrule
    Baseline+STIM+\({\cal{L}}_v\)             & 71.1 & 83.2 & 92.4 & 96.1\\
    Baseline+RRU+STIM+\({\cal{L}}_v\)             & 72.2 & 83.9 & 93.4 & 95.8\\
    Baseline+RRU+STIM+\({\cal{L}}_v\)+\({\cal{L}}_p\)           & 72.7 & 84.4 & 93.2 & 96.3\\
    \bottomrule
  \end{tabular}
  \label{mars_rru}
\end{table}

\section{Experiments}
In this section, we evaluate our method on three public video datasets for person re-identification including iLIDS-VID \cite{wang2014person}, PRID-2011 \cite{hirzer2011person} and MARS \cite{zheng2016mars}.
We first introduce the experiment setting in Sec. 4.1 and 4.2. Then, we make a ablation study on the effectiveness of each component of our method in Sec. 4.3. After that, in Sec. 4.4, we compare our approach with the state-of-the-art video-based person re-identification methods.
\subsection{Datasets and Protocols}
\textbf{iLIDS-VID} consists of 600 video sequences of 300 persons. For each person, there are two video sequences observed from two non-overlapping cameras views at an airport arrival hall.
The video sequences range in length from 23 to 192 frames with an average length of 73.
The challenging factors on this dataset include blur, occlusion and large variations in pose, viewpoints and illumination.

\textbf{PRID-2011} includes 400 video sequences of 200 identities captured by two camera views.
The length for each video sequence varies from 5 to 675 frames with an average length of 100.
Following \cite{yan2016person,zhou2017see,liu2017quality}, video sequences with more than 21 frames are used.
Compared with iLIDS-VID, PRID-2011 is relatively less challenging because of relatively simple backgrounds and rare occlusions.

\textbf{MARS} is the largest video-based person re-identification dataset, which consists of 1261 different pedestrians and around 20,000 video sequences.
Those video sequences are generated by DPM detector and GMMCP tracker, which make MARS more realistic and challenging.

Following \cite{zheng2016mars}, the partition for training and testing set in MARS dataset is given. The results are reported in terms of Cumulative Matching Characteristic (CMC) table
and mean average precision (mAP).
Following \cite{mclaughlin2016recurrent,liu2017quality}, iLIDS-VID and PRID-2011 datasets are randomly split into two sets with the same number of pedestrians for training and testing.
For testing set, video sequences from one camera view are used as probe set, while video sequences from another camera view are used as gallery set.
We use the average CMC table over 10 trials with different train/test splits to evaluate the performance of different methods on these two datasets.

\subsection{Implementation Details}
The Inception-v3 model \cite{szegedy2016rethinking} is first pretrained on the ImageNet dataset.
Each input video frame is resized to \(299 \times 299\) pixels.
It is notable that we don't apply any data augmentation strategy.
During training process, we set \(N = 10\), \(K = 2\), \(T = 8\) and \(m = 0.4\).
The dropout rate in classifier block is set to 0.5.
Network is updated by stochastic gradient descent algorithm with a learning rate of 0.01, weight decay of \(5 \times 10^{-4}\) and nesterov momentum of 0.9.
For the pretrained layers, the learning rate is set to \(0.1 \times\) of the base learning rate.
During test process, we extract the feature vector \(\textbf{f}_{i,k}\) using all frames of one video sequence \(\textbf{V}_{i,k}\) and compute the cosine distance with feature vectors of other video sequences.
The code will be made publicly available\footnote{https://github.com/yolomax/rru-reid}.
\subsection{Ablation Study of the Proposed Method}

The baseline approach contains only the Inception-v3 model trained by \({\cal{L}}_c\).
We use average pooling to generate the final representation.
As shown in Table~\ref{model_ablation}, after applying video-level ranking constraint \({\cal{L}}_v\), the rank-1 accuracy of the baseline approach is improved by 14.5\% on iLIDS-VID and 3.6\% on PRID-2011.
This shows that it is very effective to introduce the video-level ranking constraint for the networks.

\textbf{Analysis on STIM.}
When we use a LSTM with a hidden size of 256 directly to integrate temporal information, the performances are even worse than average pooling approach on both datasets because of the noise in raw features and the neglect of spatial information.
If we replace LSTM with STIM, the rank-1 accuracy is improved by a large margin, which means that STIM can better integrate spatial-temporal information.
Meanwhile, it also indicates that it is necessary to preserve the spatial resolution when mining temporal information, which makes STIM more robust to noise than LSTM.

\begin{table*}[t]
  \caption{Performance comparison with the state-of-the-art methods on iLIDS-VID, PRID-2011 and MARS datasets. The CMC scores (\%) at rank 1, 5, 20 are reported. For MARS, the mAP results are also compared.}
  \centering
  \begin{tabular}{lcccccccccc}
    \toprule
    \multirow{2}{*}{Method} & \multicolumn{3}{c}{iLIDS-VID} & \multicolumn{3}{c}{PRID-2011}  & \multicolumn{4}{c}{MARS} \\
    \cmidrule(r){2-4} \cmidrule(r){5-7} \cmidrule(r){8-11}
    & R1 & R5 & R20 & R1 & R5 & R20 & mAP & R1 & R5 & R20\\
    \midrule
    STA \cite{liu2015spatio} & 44.3 & 71.7 & 91.7 & 64.1 & 87.3 & 92.0 & - & - & - & - \\
    TDL \cite{you2016top} & 56.3 & 87.6 & 98.3 & 56.7 & 80.0 & 93.6 & - & - & - & - \\
    DVDL \cite{karanam2015person} & 25.9 & 48.2 & 68.9 & 40.6 & 69.7 & 85.6 & - & - & - & - \\
    RNN \cite{mclaughlin2016recurrent} & 58.0 & 84.0 & 96.0 & 70.0 & 90.0 & 97.0 & - & - & - & - \\
    CNN+XQDA \cite{zheng2016mars} & 53.0 & 81.4 & 95.1 & 77.3 & 93.5 & 99.3 & 49.3 & 68.3 & 82.6 & 89.4 \\
    SeeForest \cite{zhou2017see} & 55.2 & 86.5 & 97.0 & 79.4 & 94.4 & 99.3 & 50.7 & 70.6 & 90.0 & \textbf{97.6} \\
    TSSCNN \cite{chung2017two} & 60.0 & 86.0 & 97.0 & 78.0 & 94.0 & 99.0 & - & - & - & - \\
    ASTPN \cite{xu2017jointly} & 62.0 & 86.0 & 98.0 & 77.0 & 95.0 & 99.0 & - & 44.0 & 70.0 & 81.0 \\
    QAN \cite{liu2017quality} & 68.0 & 86.8 & 97.4 & 90.3 & 98.2 & \textbf{100.0} & - & - & - & - \\
    RQEN \cite{song2017region} & 76.1 & 92.9 & 99.3 & 92.4 & \textbf{98.8} & 100.0 & 51.7 & 73.7 & 84.9 & 91.6 \\
    DRSTA \cite{li2018diversity} & 80.2 & - & - & \textbf{93.2} & - & - & 65.8 & 82.3 & - & - \\
    \midrule
    Baseline & 57.7 & 78.9 & 91.3 & 84.4 & 96.2 & 99.5 & 64.7 & 81.7 & 91.8 & 96.2\\
    Ours & \textbf{84.3} & \textbf{96.8} & \textbf{99.5} & 92.7 & \textbf{98.8} & 99.8 & \textbf{72.7} & \textbf{84.4} & \textbf{93.2} & 96.3\\
    \bottomrule
  \end{tabular}
  \label{state_art}
\end{table*}

\textbf{Analysis on RRU.}
When we apply RRU to \emph{Baseline+\({\cal{L}}_v\)}, the rank-1 accuracy is improved by 2.8\% on iLIDS-VID and 3.3\% on PRID-2011.
We show some examples of the feature maps before and after being refined by RRU in Fig.~3.
For the left video sequence fragment, due to the severe blur, the responses to the legs of the raw feature maps in the second frame and third frame are almost lost as shown in the white boxes.
After refinement by RRU, the responses are well recovered.
For the middle video sequence fragment, the model suffers from background clutter.
As shown in the gold box in the second frame, the background part has noisy responses.
After the process of RRU, the noisy responses are well removed.
Meanwhile, the drifting responses of upper body in raw feature maps of second frame and the nearly missing responses in the white box of the fourth frame are well corrected.
For the right video sequence fragment, the hand in the fourth frame is occluded by the sign.
After refinement by RRU, the responses are well recovered, and the almost lost responses in the first frame are also well recovered.
Those examples verify the denoising and recovering capabilities of RRU.

We evaluated the effects of update gate model \(g\) with different component setting. As shown in Table~\ref{model_ablation}, the combination of \(\textbf{Z}_{s}\) and \(\textbf{Z}_{c}\) achieves better performances than using them alone.
We also investigate the effect of the input information for update gate model \(g\), which is shown in Table~\ref{model_ablation}.
When we remove the motion information \([\textbf{X}_{i, k, t}- \textbf{X}_{i, k, t-1}]\) and only keep the appearance differences \([\textbf{X}_{i, k, t}- \textbf{S}_{i, k, t-1}]\), the performance declines on both datasets.
This indicates the motion information is useful for \(g\) to evaluate the quality of feature maps.
When we directly use the concatenated feature maps \([\textbf{X}_{i, k, t}, \textbf{S}_{i, k, t-1}]\) as input without explicitly indicating the appearance differences and motion information, the performance also drops.
This indicates that it is more efficient to provide the priori information about appearance differences and motion information than to let \(g\) directly learn how to integrate the information from the \(\textbf{S}_{i, k, t-1}\) and \(\textbf{X}_{i, k, t}\).

When we insert RRU before LSTM, the rank-1 accuracy is improved by 0.8\% on iLIDS-VID and 2.9\% on PRID-2011.
When RRU is inserted before STIM, RRU boosts the rank-1 accuracy by 3.2\% on iLIDS-VID and 0.4\% on PRID-2011.
Specifically, when collaborated with RRU, STIM achieves greater performance improvement than LSTM on challenging iLIDS-VID dataset, since STIM makes better utilization of the refined appearance representation generated by RRU, while LSTM ignores the spatial information.
After applying part-level ranking constraint on RRU, RRU further boosts the rank-1 accuracy of STIM by 3.8\% on iLIDS-VID and 1.2\% on PRID-2011.
The video sequences of MARS dataset \cite{zheng2016mars} are generated by DPM detector and GMMCP tracker, which are not well-aligned and make it more challenging than other datasets. As shown in Table~\ref{mars_rru}, RRU and \({\cal{L}}_p\) still boost the accuracy. RRU refines feature maps instead of raw RGB images. The large receptive fields of neurons in the backbone model mitigate the impact of imperfect pedestrian detection bounding box.
We can conclude that RRU is well compatible with other models.
Meanwhile, the combination of video-level ranking constraint and part-level ranking constraint is important for further improving the learning capability of networks.

\subsection{Comparison to the State-of-the-Art methods}

We compare the performance of our approach with other state-of-the-art methods on three datasets in Table~\ref{state_art}.
Our approach achieves the best performance on iLIDS-VID and MARS for rank-1 accuracy and mAP accuracy.
Compared with DRSTA, the improvements achieved by our approach are 4.1\% and 2.1\% for rank-1 accuracy on iLIDS-VID and MARS, respectively.
Specifically, for MARS, our approach exceeds DRSTA by 6.9\% for mAP accuracy.
On PRID-2011 dataset, our approach is only slightly lower than DRSTA in terms of rank-1 accuracy.
As discussed before, PRID-2011 is relatively less challenging because of relatively simple background and rare occlusion, which makes the capability of our model fail to be fully expressed.


\section{Conclusions}
In this paper, we propose a new network architecture for video-based person re-identification,
which consists of two key modules, \emph{i.e.,} refining recurrent unit (RRU) and spatial-temporal clues integration module (STIM).
The former refines the feature activations with a recurrent paradigm, while the latter integrates the abundant spatial-temporal information.
Those modules are collaboratively optimized with a multi-level training objective.
Extensive experiments on three popular benchmark video datasets demonstrate the effectiveness of our approach.

\section{ Acknowledgments}
This work was supported in part to Prof. Houqiang Li by 973 Pro-gram under contract No. 2015CB351803 and NSFC under contract No. 61390514, and in part to Dr. Wengang Zhou by NSFC under Contract 61822208 and Contract 61632019, the Fundamental Research Funds for the Central Universities, the Young Elite Scientists Sponsorship Program by the CAST under Grant 2016QNRC001, and the Youth Innovation Promotion Association CAS under Grant 2018497.

\bibliographystyle{aaai}
\bibliography{formatting-instructions-latex-2019}
\end{document}